\RequirePackage{fix-cm}
\documentclass[11pt,a4paper]{article}  

\usepackage{type1cm}        % activate if the above 3 fonts are
\usepackage{makecell}
\usepackage{placeins}
\usepackage{hyperref}
\usepackage{makeidx}         % allows index generation
\usepackage{graphicx}        % standard LaTeX graphics tool
\usepackage{multicol}        % used for the two-column index
\usepackage[bottom]{footmisc}% places footnotes at page bottom
\usepackage{multirow}

\usepackage{newtxtext}       % 
\usepackage{newtxmath}       % selects Times Roman as basic font
\usepackage{caption}
\usepackage[normalem]{ulem}

\makeindex             % used for the subject index
\begin{document}

\title{A deep learning approach for detection and localization of leaf anomalies}

\author{Davide Calabr\`o$^1$, Massimiliano Lupo Pasini$^2$, Nicola Ferro$^3$, Simona Perotto$^3$}
\date{}
\maketitle

\begin{center}
{\small
$^1$
Dipartimento di Matematica\\
Politecnico di Milano\\
Piazza L. da Vinci, 32, I-20133 Milano, Italy\\
{\tt davide.calabro@mail.polimi.it}
\\[3mm]
$^2$
Oak Ridge National Laboratory\\
1 Bethel Valley Road, Oak Ridge, TN, USA, 37831\\ {\tt lupopasinim@ornl.gov}
\\[3mm]
$^3$
MOX -- %Modellistica e Calcolo Scientifico\\
Dipartimento di Matematica\\
Politecnico di Milano\\
Piazza L. da Vinci, 32, I-20133 Milano, Italy\\
{\tt \{nicola.ferro, simona.perotto\}@polimi.it}
}
\end{center}
\date{}
\maketitle

\begin{abstract}
The detection and localization of possible diseases in crops are usually automated by resorting to supervised deep learning approaches. In this work, we tackle these goals with unsupervised models, by applying three different types of autoencoders to a specific open-source dataset of healthy and unhealthy pepper and cherry leaf images. CAE, CVAE and VQ-VAE autoencoders are deployed to screen unlabeled images of such a dataset, and compared in terms of image reconstruction, anomaly removal, detection and localization. The vector-quantized variational architecture turns out to be the best performing one with respect to all these targets.
\\[0.5cm]
{\footnotesize This manuscript has been authored in part by UT-Battelle, LLC, under contract DE-AC05-00OR22725 with the US Department of Energy (DOE). The US government retains and the publisher, by accepting the article for publication, acknowledges that the US government retains a nonexclusive, paid-up, irrevocable, worldwide license to publish or reproduce the published form of this manuscript, or allow others to do so, for US government purposes. DOE will provide public access to these results of federally sponsored research in accordance with the DOE Public Access Plan (\url{http://energy.gov/downloads/doe-public-access-plan})}
\end{abstract}

\section{Introduction}

In the last decades, the agricultural sector has been developing new technologies to maximize the efficiency of available soil resources in order to tackle several issues. Instances are the increasing demand for food due to the growth of the world population~\cite{onu_food_2018}, as well as the impact of practices which are detrimental for the ecosystem~\cite{dale07}. In these contexts, precision agriculture has recently attracted a lot of interest since playing a significant role in the development of advanced techniques that optimize the soil productivity in a sustainable way \cite{zhang02,messick17}. The general goal is to preserve the stability of the ecosystem while fostering the reuse of the soil for future produce. The strong interest in this new way of conceiving agriculture justifies the spread of innovative start-ups, of services devoted to eco-friendly practices, and of software solutions which allow farmers to accurately estimate yields on a simple smartphone or tablet (see, for instance,~\cite{cleverfarm,dynacrop,pixofarm,plantix}). In particular, the 
availability of higher-quality measurements, offered by advanced in field-sensors as well as by satellite or drone data, supported the proposal of breakthrough software solutions using modern deep learning algorithms~\cite{kamilaris17,liakos_machine_2018,whelan18,vuran18,sharma20}.

In this paper, we focus on the detection of possible diseases in crops. Anomaly detection in plants represents a pivotal procedure in agriculture since an early detection of the disease enables a timely intervention to prevent the anomaly from spreading to the rest of the plant. Additionally, a precise disease localization allows confining the use of pesticides and other treatments to small strategically selected areas of the plant. This minimizes the negative impact of aggressive chemicals on the whole surrounding ecosystem. 
\\
Traditionally, the detection of the diseases is carried out manually. This is a time-consuming task and turns out to be expensive in terms of human resources. 
The development of advanced technologies as well as modern devices has recently offered the possibility to perform disease detection in a more efficient and affordable way by automating the process. For instance, computer vision turns out to be instrumental in analyzing large quantities of images collected by drones in a short time, such as photographs of healthy and diseased leaves in crops~\cite{agrio}.
Among the several computer vision techniques, machine learning (ML) methods, with a focus on deep learning (DL) techniques, showed great potentiality in accurately discriminating healthy from unhealthy leaves through suitable classification procedures.

In the supervised learning framework, the DL models need to be fed with a large volume of training images in order to thoroughly span all classes in the dataset \cite{FERNANDEZ202277, 8999187}.
In particular, in the leaf detection context, during the training phase a supervised DL method deals with healthy and unhealthy leaf classes in order to offer an accurate classification.
This turns into a not so feasible practice due to the limited availability of images of leaves affected by diseases.
\\
The importance of the selected dataset is confirmed by several contributions in the literature. For instance, in~\cite{10.3389/fpls.2016.01419} the authors train a convolutional neural network (CNN) to identify 14 crop species and 26 diseases to conclude that only training DL models on increasingly large image datasets allows guaranteeing a reliable crop diagnosis. 
In~\cite{10.3389/fpls.2019.00941}, the authors investigate the sensitivity characterizing the supervised learning of CNN models to the accuracy of the data. In particular, it is highlighted that the lack of conformity across different data samples can severely impact the generalizability of CNNs in accurately classifying diseased and healthy leaves.

The idea behind unsupervised DL procedures for anomaly detection is to carry out the training only on the healthy part of the dataset in order to accurately reproduce non-anomalous samples, while minimizing the risk of false negatives~\cite{pinaya_unsupervised_2021}. Once validated, the DL model is deployed for anomaly detection on unlabeled images. If the sample is characterized by a reconstruction error below a threshold, very likely it belongs to the same distribution as the training data, and it is thus classified as non-anomalous. Vice versa, if the reconstruction error is large, the data sample is very probably generated from a distribution different from the training data, and the sample is thus classified as anomalous. 
\\
DL models generally used for unsupervised anomaly detection are convolutional autoencoders (CAEs) \cite{wang_image_2020, morawski2021}, which are commonly exploited also for anomaly localization (see, e.g., \cite{https://doi.org/10.48550/arxiv.1911.08616, https://doi.org/10.48550/arxiv.2012.11113}).

As a third alternative, contributions where supervised and unsupervised approaches are combined are available in the literature. For instance, in \cite{9158218} the authors use an unsupervised CAE model to derive relevant features from images of leaves, and then feed them into support vector machine (SVM) models for classification. Similarly, in \cite{HARAKANNANAVAR2022305} unsupervised learning techniques, such as principal component analysis and K-means, are employed to recover the informative features of samples of tomato leaves. Then, the extracted characteristics are classified by resorting to supervised ML approaches, such as SVM, CNN and K-nearest neighbors.

In this work, we apply unsupervised learning CAE models to detect and localize leaf diseases. 
Up to our knowledge, this type of data is still scarcely analyzed in the literature with unsupervised DL approaches. Goal of the paper is to compare the performance of three different CAE architectures, namely standard CAEs, convolutional variational autoencoders (CVAEs), vector-quantized variational autoencoders (VQ-VAEs), on the PlantVillage dataset, which collects images of healthy and diseased leaves of several species. The three models are assessed in terms of image reconstruction and anomaly removal, detection and localization.
\\
The paper is organized as follows. Section~\ref{sec:autoencoders} details the three convolutional autoencoders. Section~\ref{sec:leaf} focuses on the workflow adopted for the anomaly detection, while providing the main features of the considered image samples. In Section~\ref{sec:verification}, we gather more technical information concerning the data preprocessing, the model setup, hardware and software specifics. Section~\ref{sec:discussion} discusses the reliability and the performance of the three compared architectures, with a specific emphasis both on the reconstruction of the images and on the anomaly detection/localization. Finally, some conclusions are drawn in the last section, together with possible future developments.

\section{Convolutional autoencoders}\label{sec:autoencoders}
In this section, we describe the convolutional autoencoder (CAE) architecture together with two recent improved stabilized variants, namely the convolutional variational autoencoders (CVAEs) and the vector-quantized variational autoencoders (VQ-VAEs).

\begin{figure}[h]
\centering
\includegraphics[width=\textwidth]{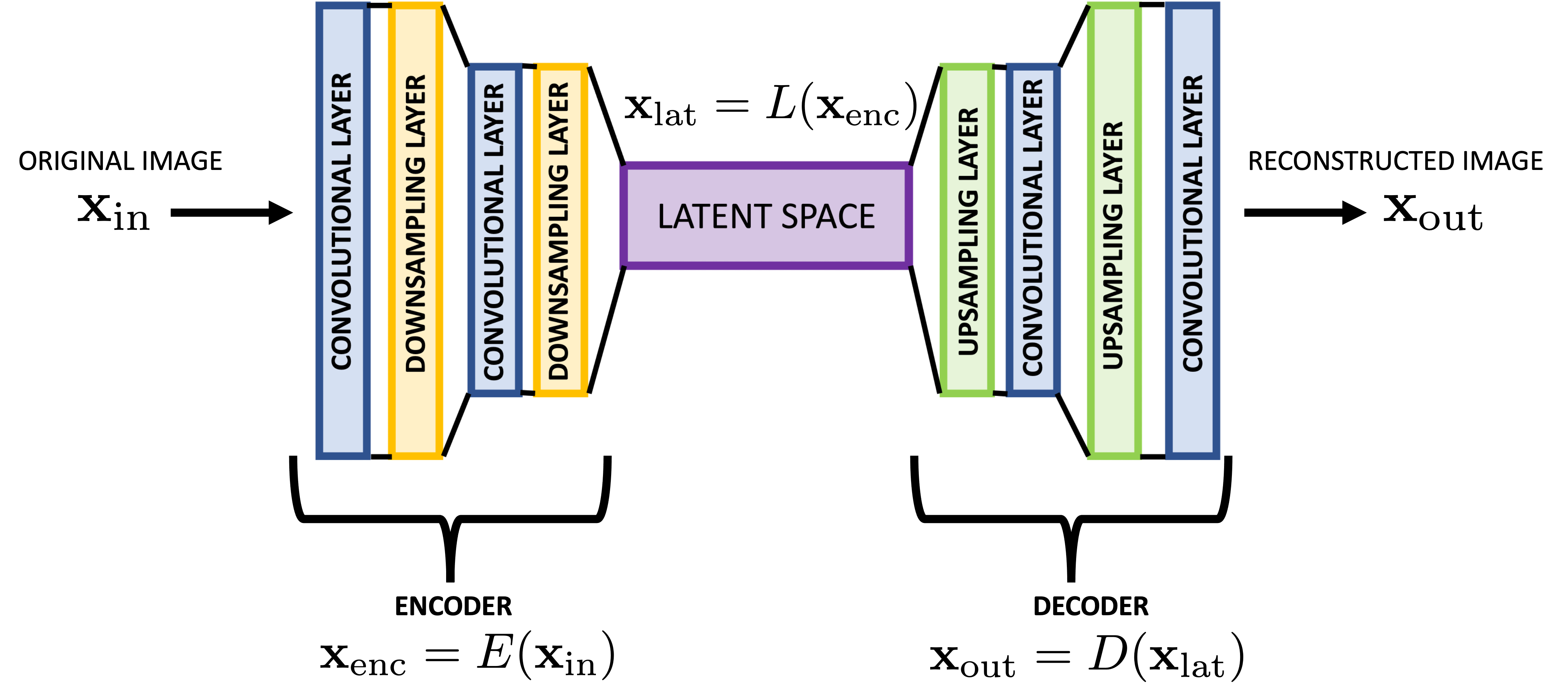}
\caption{Architecture of a convolutional autoencoder.}\label{fig:methodology_cae}
\end{figure}

CAE models can be considered as a dimensionality reduction technique, since they allow extracting essential features from image data to enable an effective compression with a minimal loss of information. The architecture of a CAE model consists of three main components: an encoder, a stack of fully connected layers (latent space), and a decoder. A schematic representation of a CAE is provided in Figure~\ref{fig:methodology_cae}.

An image is generally represented as a three-dimensional (3D) tensor of size $(H, W, C)$, where $H$ and $W$ denote the number of pixels in height and width, respectively, while $C$ is the number of channels (i.e., the depth) of the tensor.
An RGB image has three channels, thus resulting into a tensor of size $(H,W,3)$. Hereafter, we will adopt a vector in $\mathbb{R}^N$ as a representation of an RGB image equivalent to such a tensor, with $N=3N_c$ and $N_c = H W$ the number of pixels per channel.\\
An RGB image $\mathbf{x}_\text{in} \in \mathbb{R}^N$ represents the standard input to a CAE encoder, which is assembled as a convolutional neural network (CNN) architecture that alternates convolutional with downsampling (average-pooling or max-pooling) layers in order to compress the input image~\cite{10.1007/978-3-030-66151-9_17}. The encoder transforms an input RGB image into a 3D tensor of size $(\tilde{H},\tilde{W},\tilde{C})$, with $\tilde{H} < H$, $\tilde{W}< W$, and $\tilde{C}>3$. The encoder action is mathematically represented by a nonlinear operator $E$ such that $\mathbf{x}_{\text{enc}} = E(\mathbf{x}_\text{in})$, with $\mathbf{x}_{\text{enc}}$ the compressed image. 
\\
The encoded image is successively mapped into the latent space, namely $\mathbf{x}_{\text{lat}} = L(\mathbf{x}_{\text{enc}})$, with $L$ a nonlinear operator.
\\
Finally, the re-elaborated compressed image  $\mathbf{x}_{\text{lat}}$ is fed into the decoder, which restores the information lost at the highest attainable fidelity. The decoder is constructed as a CNN architecture that alternates convolutional
with upsampling layers in order to progressively re-expand the image up to the original size. The decoder is mathematically represented by a nonlinear operator $D$, which computes the final CAE output as an RGB image $\mathbf{x}_{\text{out}} = D(\mathbf{x}_\text{lat})$.
\\
The nonlinear operators $E$, $L$ and $D$ are selected as regression models. We denote by $\mathbf{w}_1 \in \mathbb{R}^{N_{\text{1}}}$, $\mathbf{w}_2 \in \mathbb{R}^{N_{\text{2}}}$, and $\mathbf{w}_3 \in \mathbb{R}^{N_{\text{1}}}$ the vectors gathering the corresponding regression coefficients. As a consequence, the generic actions in Figure~\ref{fig:methodology_cae} can be particularized as
$$\mathbf{x}_{\text{enc}} = E_{\mathbf{w}_1}(\mathbf{x}_\text{in}),\quad
\mathbf{x}_{\text{lat}} = L_{\mathbf{w}_2}(\mathbf{x}_{\text{enc}}),\quad
\mathbf{x}_{\text{out}} = D_{\mathbf{w}_3}(\mathbf{x}_\text{lat}),
$$ 
with 
$$
E_{\mathbf{w}_1}:\mathbb{R}^{N_1}\times 
\mathbb{R}^{N} \rightarrow \mathbb{R}^{N_{\text{lat}}},
\ 
L_{\mathbf{w}_2}: \mathbb{R}^{N_2}\times  \mathbb{R}^{N_{\text{lat}}} \rightarrow \mathbb{R}^{N_{\text{lat}}},
\ 
D_{\mathbf{w}_3}:\mathbb{R}^{N_1}\times \mathbb{R}^{N_{\text{lat}}} \rightarrow \mathbb{R}^{N},
$$
where $N_{\text{lat}} = \tilde{H} \tilde{W} \tilde{C} < N$.
The overall action of the CAE can be expressed in a compact form by introducing the vector $\mathbf{w} = \big[ \mathbf{w}_1, \mathbf{w}_2, \mathbf{w}_3 \big]^T\in \mathbb{R}^{N_{\text{param}}}$ collecting all the regression coefficients, with $N_{\text{param}} = 2 N_{\text{1}} + N_{\text{2}}$, and the nonlinear operator $C_\mathbf{w}:\mathbb{R}^{N_{\text{param}}}\times\mathbb{R}^{N}\rightarrow \mathbb{R}^N$ defined by 
\begin{equation}
    C_\mathbf{w}(\mathbf{x}_{\text{in}}) = D_{\mathbf{w}_3}(L_{\mathbf{w}_2}(E_{\mathbf{w}_1}(\mathbf{x}_{\text{in}}))) = \mathbf{x}_{\text{out}}.
\end{equation}

The training of the CAE over a set of input images $\{ \mathbf{x}_{\text{in}}\}$ aims at compressing and successively decompressing each input image by producing an approximation $\mathbf{x}_{\text{out}}$, with a minimal loss of information with respect to $\mathbf{x}_{\text{in}}$. 
To this goal, the loss function, coinciding with the mean-square error
\begin{equation}\label{MSE}
    \mathcal{L}_{\rm CAE}(\mathbf{x}_{\text{in}}, \mathbf{x}_{\text{out}}; \mathbf{w}) = {\rm MSE}(\mathbf{x}_{\text{in}}, \mathbf{x}_{\text{out}}; \mathbf{w}) = \mathbb{E}\big[ \lVert \mathbf{x}_{\text{in}} - \mathbf{x}_{\text{out}} \rVert^2 \big],
\end{equation}
is minimized by varying $\mathbf{w} \in \mathbb R^{N_{\text{param}}}$, with $\mathbb{E}[\cdot]$ the expected value and
$\| \cdot \|$ the Euclidean norm. In particular, such a minimization is carried out over successive batches of data and by using batched stochastic optimization with automatic differentiation \cite{Robbins_1951}.

\subsection{Convolutional variational autoencoders}

\begin{figure}[h]
\centering
\includegraphics[width=\textwidth]{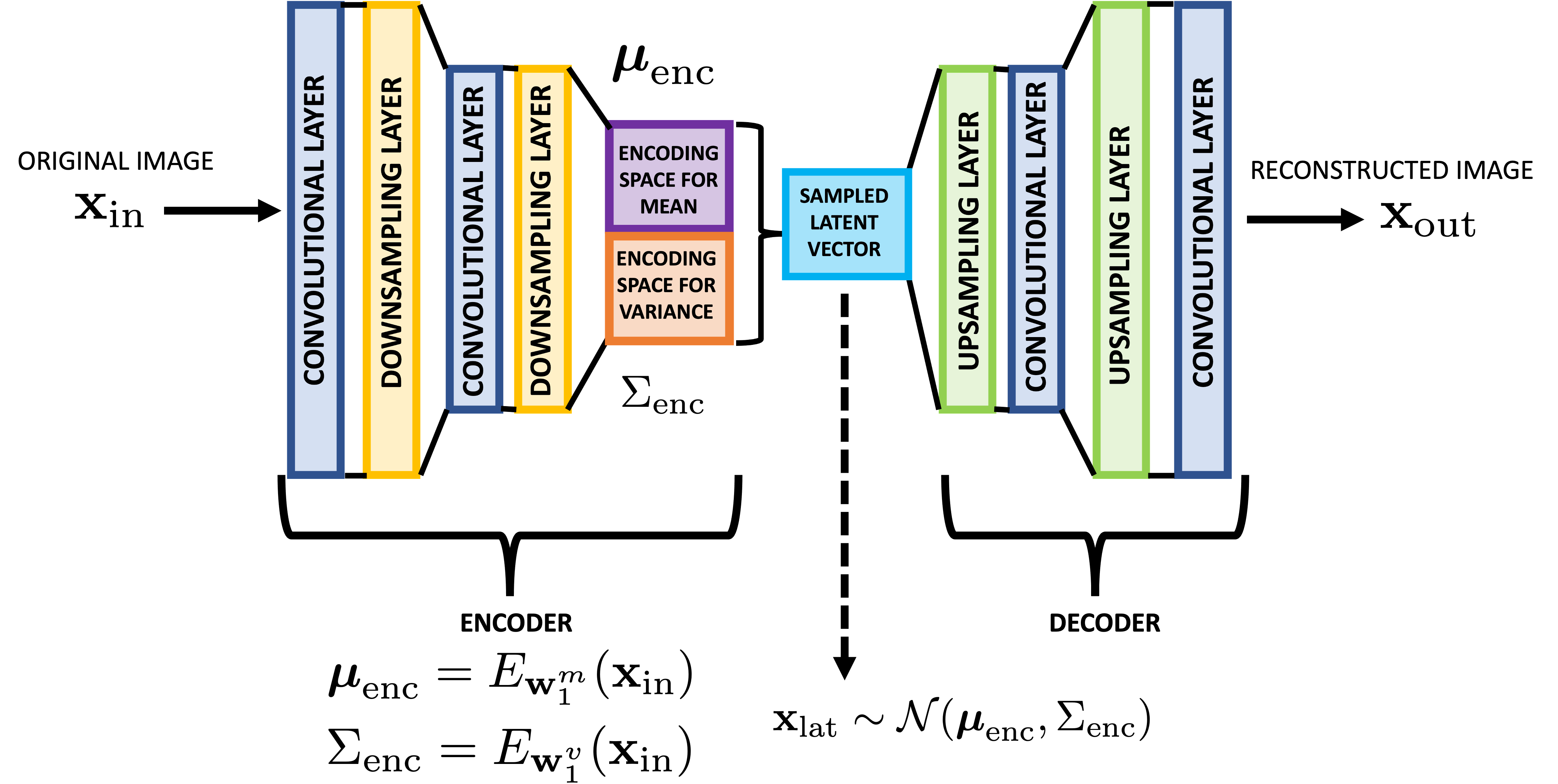}
\caption{Architecture of a convolutional variational autoencoder.}\label{fig:methodology_cvae}
\end{figure}

CAEs may suffer from overfitting during the training. CVAE models offer an improvement of convolutional autoencoders, by properly regularizing such a phase.
In particular, the peculiar difference between CAE and CVAE models lies in the definition of the latent space. Standard CAE models encode the input as a deterministic vector in the latent space and feed the decoder with an analogous type of data. On the contrary, CVAEs encode the input according to a distribution $\mathcal D$ of a continuous variable defined over the latent space, and the decoder is fed with a sample from that distribution. In general, we choose $\mathcal D$ as a multi-variate Gaussian distribution characterized by a mean, $\boldsymbol{\mu}_{\text{enc}}$, and by 
a diagonal variance matrix, whose entries are gathered in the vector $\Sigma_{\text{enc}}$. According to a CVAE architecture, vectors $\boldsymbol{\mu}_{\text{enc}}$ and $\Sigma_{\text{enc}}$ are computed by two separate encoders (see Figure \ref{fig:methodology_cvae}).

To provide a mathematical formalization of such a process, we introduce vectors $\mathbf{w}_{1}^m$, $\mathbf{w}_1^v\in \mathbb{R}^{N_1}$ collecting the coefficients associated with the regression encoders $E_{\mathbf{w}_1^m}$, $E_{\mathbf{w}_1^v}:\mathbb{R}^{N_1\times N}\rightarrow \mathbb{R}^{N_{\text{lat}}}$, for the mean and the variance, respectively, such that
\begin{equation}
\boldsymbol{\mu}_{\text{enc}} = E_{\mathbf{w}_1^m}(\mathbf{x}_{\text{in}}), \quad 
\Sigma_{\text{enc}} = E_{\mathbf{w}_1^v}(\mathbf{x}_{\text{in}}),
\end{equation}
with $\mathbf{x}_{\text{in}} \in \mathbb{R}^N$ the input image, and $N_{\rm lat} = \tilde{H} \tilde{W} \tilde{C} < N$, the dimension of the latent space. 
\\
The output of the encoder, $\boldsymbol{\mu}_{\text{enc}}$ and $\Sigma_{\text{enc}}$, are fed into the latent space that computes the latent vector
\begin{equation}
    \mathbf{x}_{\text{lat}} = L_\mathcal{N}(\boldsymbol{\mu}_{\text{enc}}, \Sigma_{\text{enc}})
\end{equation}
by randomly sampling the multi-variate Gaussian distribution $\mathcal{N}(\boldsymbol{\mu}_{\text{enc}}, \Sigma_{\text{enc}})$, with $L_\mathcal{N}: \mathbb{R}^{N_{\text{lat}}} \times \mathbb{R}^{N_{\text{lat}}} \rightarrow \mathbb{R}^{N_{\text{lat}}}$ the operator associated with the latent space.
\\
Finally, the sample $\mathbf{x}_{\text{lat}}$ is passed to the decoder, represented by the regression operator $D_{\mathbf{w}_3}:\mathbb{R}^{N_1\times N_{\text{lat}}}\rightarrow \mathbb{R}^{N}$, to generate the reconstructed image
\begin{equation}
    \mathbf{x}_{\text{out}} = D_{\mathbf{w}_3}(\mathbf{x}_{\text{lat}}),
\end{equation}
with $\mathbf{w}_3$ the vector of the regression coefficients characterizing the decoder.

The overall action of the CVAE model can be expressed in a compact form by introducing the vector $\mathbf{w} = \big[ \mathbf{w}_1^m, \mathbf{w}_1^v, \mathbf{w}_3 \big]^T\in \mathbb{R}^{N_{\text{param}}}$ collecting all the regression coefficients, with $N_{\text{param}} = 3 N_{\text{1}}$, and the nonlinear operator $C_\mathbf{w}:\mathbb{R}^{N_{\text{param}}}\times\mathbb{R}^{N}\rightarrow \mathbb{R}^N$ defined by 
\begin{equation}
    C_\mathbf{w}(\mathbf{x}_{\text{in}}) = D_{\mathbf{w}_3}(
    L_\mathcal{N}(E_{\mathbf{w}_1^m}(\mathbf{x}_{\text{in}}), E_{\mathbf{w}_1^v}(\mathbf{x}_{\text{in}}))) = \mathbf{x}_{\text{out}}.
\end{equation}

The involvement of a continuous variable in the CVAE model leads to add a term related to the multi-variate Gaussian distribution $\mathcal{N}(\boldsymbol{\mu}_{\text{enc}}, \Sigma_{\text{enc}})$ to the loss function in \eqref{MSE}. Thus, the training is performed by minimizing the new loss function
\begin{equation}\label{KL}
    \mathcal{L}_{\rm CVAE}(\mathbf{x}_{\text{in}}, \mathbf{x}_{\text{out}}; \mathbf{w}) = \mathbb{E}\big[ \lVert \mathbf{x}_{\text{in}} - \mathbf{x}_{\text{out}} \rVert^2 \big] + D_{\rm KL}\big(\mathcal{N}(\mu_{\text{enc}}, \Sigma_{\text{enc}}), \mathcal{N}(\mathbf{0},I)\big),
\end{equation}
with $D_{\rm KL}$ the so-called Kullback-Leibler (KL) divergence~\cite{kullback_information_1951}. We observe that the new contribution in \eqref{KL} plays the role of a regularizing term by measuring the discrepancy between the distribution returned by the encoder, $\mathcal{N}(\mu_{\text{enc}}, \Sigma_{\text{enc}})$, and a standard Gaussian distribution, $\mathcal{N}(\mathbf{0}, I)$.
For more details about the CVAE architecture, we refer the reader to \cite{Kingma_2019}. 

When the training data is limited, using a regularization defined in a continuous space as in \eqref{KL} may introduce a strong bias, which results into a severe underfitting with significant performance deterioration. 
To overcome this issue, new autoencoding architectures have been developed, as discussed in the following.

\subsection{Vector-quantized variational autoencoders}
\begin{figure}[h]
\centering
\includegraphics[width=0.7\textwidth]{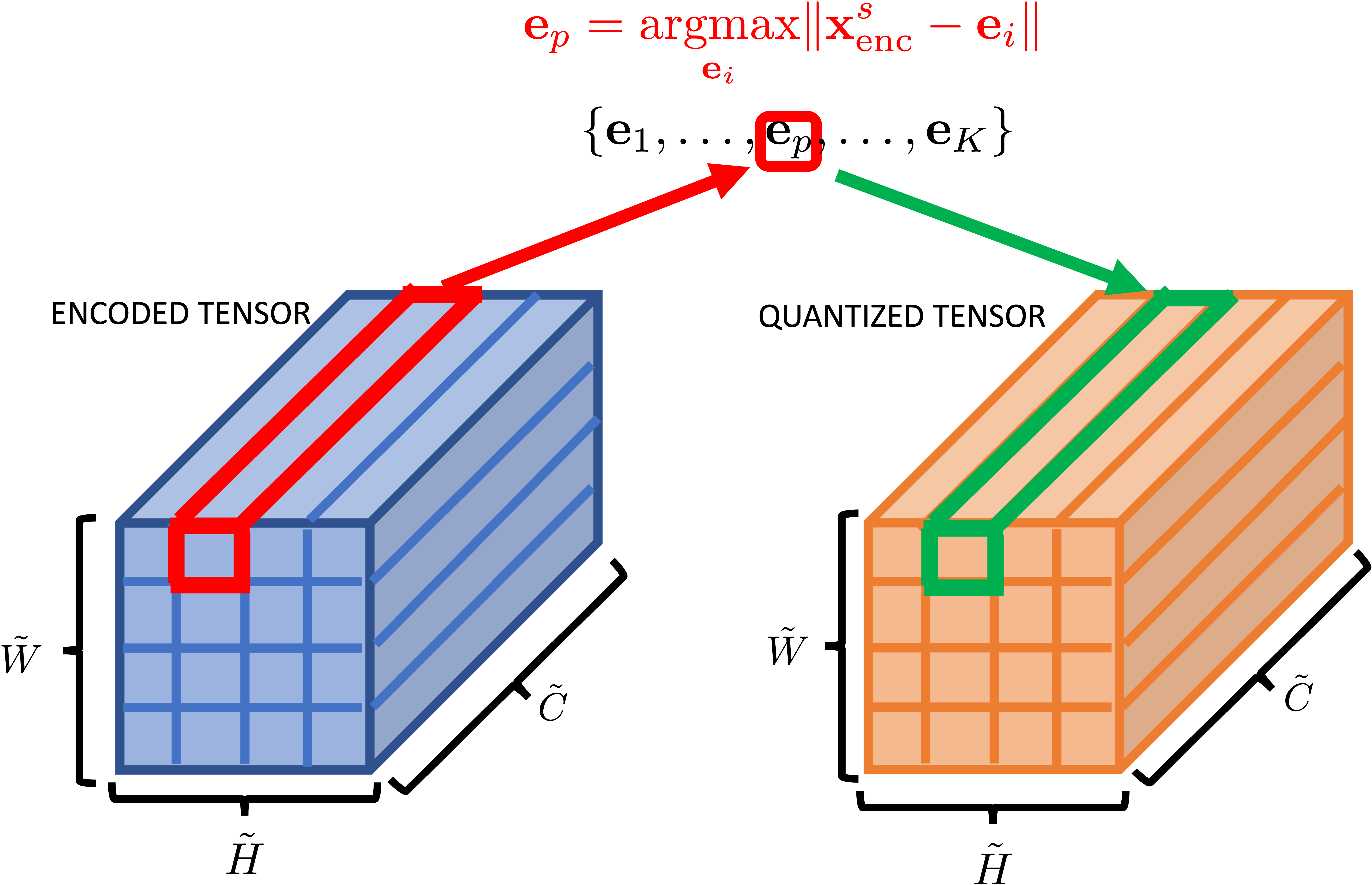}
\caption{Illustration of the quantization of the encoded vector in terms of the codebook.}\label{fig:methodology_quantization}
\end{figure}

\begin{figure}[h]
\centering
\includegraphics[width=\textwidth]{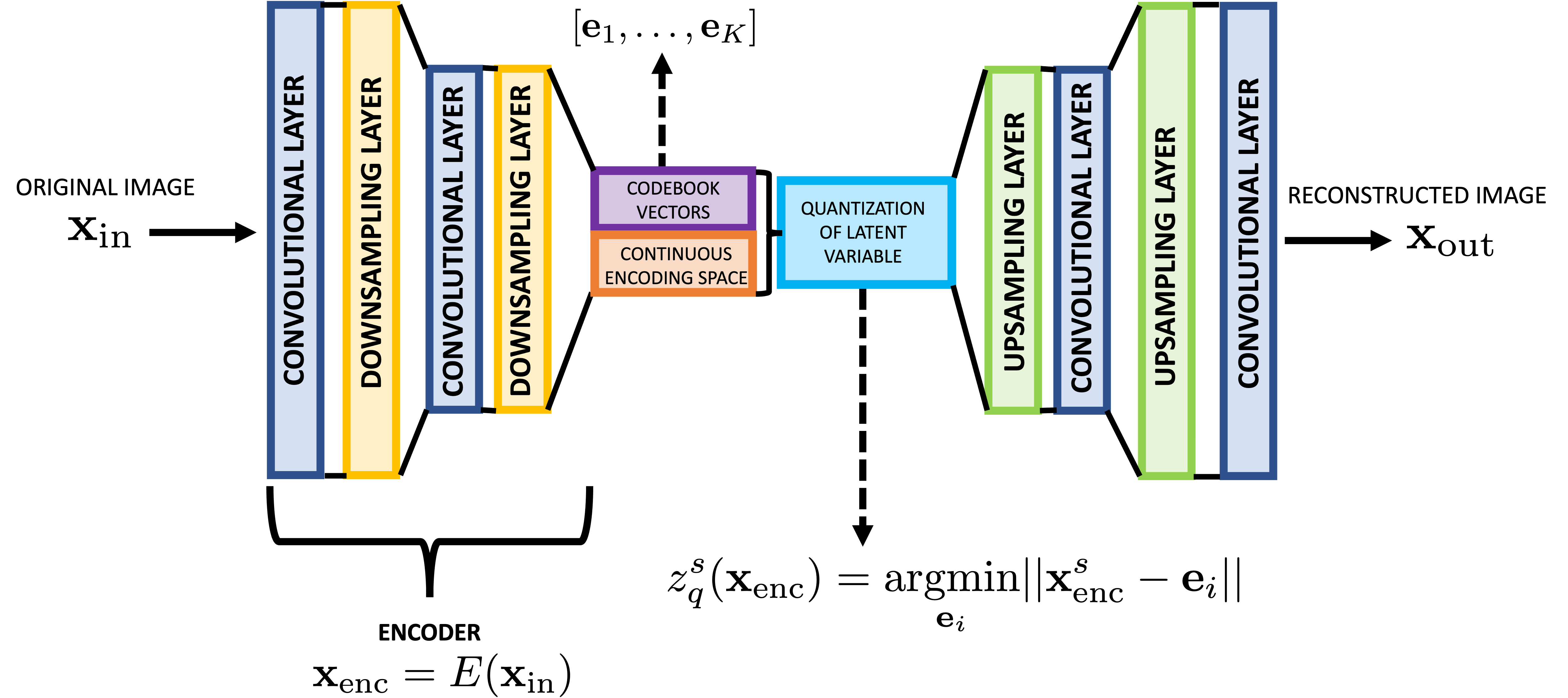}
\caption{Architecture of a vector-quantized variational autoencoder.}\label{fig:methodology_vqvae}
\end{figure}

VQ-VAE models offer a solution to the biasing effect triggered when the training data is small, by replacing the continuous regularization in \eqref{KL} with a new term defined in a discrete space.
In particular, this space is characterized by a set of $K$ learnable vectors, $\{\mathbf{e}_i\}_{i=1}^K$, called codebook, with $\mathbf{e}_i \in \mathbb{R}^{\tilde{C}}$. The encoded vector $\mathbf{x}_{\text{enc}} =  E_{{\bf w}_1}(\mathbf{x}_{\text{in}}) \in \mathbb{R}^{N_{\rm lat}}$, with $N_{\rm lat}=\tilde{H} \tilde{W} \tilde{C}$, 
is remapped into a 3D tensor of size $(\tilde{H}, \tilde{W}, \tilde{C})$ and then cut into longitudinal threads following the third dimension, creating a total of $\tilde{H} \tilde{W}$ fibers, $\mathbf{x}^s_{\text{enc}}\in \mathbb{R}^{\tilde{C}}$, for $s=1,\ldots, \tilde{H} \tilde{W}$. %(see Figure~\ref{fig:methodology_quantization}, left).
For each fiber, the vector in the codebook $\{\mathbf{e}_i\}_{i=1}^K$ which is closest to $\mathbf{x}^s_{\text{enc}}$ is detected. This phase can be formalized by the minimization
\begin{equation}
    z_q^s(\mathbf{x}_{\rm enc}) = \mathbf{e}_p = \underset{\mathbf{e}_i}{\operatorname{argmin}}  ||\mathbf{x}^s_{\text{enc}} - \mathbf{e}_i|| \quad s = 1, \ldots, \tilde{H} \tilde{W},
    \label{codebook_learning}
\end{equation}
where $z_q^s(\mathbf{x}_{\rm enc}) \in \mathbb{R}^{\tilde{C}}$ coincides with the quantized representation of the thread $\mathbf{x}^s_{\text{enc}}$ (see Figure \ref{fig:methodology_quantization}). The threads $z_q^s(\mathbf{x}_{\rm enc})$ are eventually aggregated to yield the vector ${\bf z}_q^{\rm lat} \in \mathbb{R}^{N_\text{lat}}$. \\
We mathematically represent all the operations employed to convert the encoded vector $\mathbf{x}_{\text{enc}}$ into ${\bf z}_q^{\rm lat}$ by the nonlinear operator $L_{{\bf q}}:\mathbb{R}^{N_2}\times \mathbb{R}^{N_\text{lat}}\rightarrow \mathbb{R}^{N_\text{lat}}$, with $\mathbf{q} \in \mathbb{R}^{N_2}$ the vector collecting the corresponding regression coefficients.
Finally, vector ${\bf z}_q^{\rm lat}$ is passed to the decoder for the final image reconstruction.
Thus, the overall action of the VQ-VAE model can be expressed in a compact form by introducing the vector $\mathbf{w} = \big[ \mathbf{w}_1, \mathbf{q}, \mathbf{w}_3 \big]^T\in \mathbb{R}^{N_{\text{param}}}$ collecting all the regression coefficients, with $N_{\text{param}} = 2 N_{\text{1}} + N_{\text{2}}$, and the nonlinear operator $C_\mathbf{w}:\mathbb{R}^{N_{\text{param}}}\times\mathbb{R}^{N}\rightarrow \mathbb{R}^N$, such that 
\begin{equation}
    C_\mathbf{w}(\mathbf{x}_{\text{in}}) = D_{\mathbf{w}_3}(
    L_{\mathbf{q}}(E_{\mathbf{w}_1}(\mathbf{x}_{\text{in}}))) = \mathbf{x}_{\text{out}}.
\end{equation}
\\
An illustration of the VQ-VAE architecture is provided in Figure \ref{fig:methodology_vqvae}. 
%(see Figure~\ref{fig:methodology_quantization}).

The training of a VQ-VAE model includes the codebook as well as the model learning through a backpropagation process. Similarly to the loss function of CVAE, the training of the VQ-VAE model is carried out by iteratively minimizing a loss function consisting of a reconstruction error enriched by a regularizing term. In particular, the regularization coincides with the codebook learning, which is performed by the minimization in \eqref{codebook_learning}.
\\
In particular, the codebook learning is characterized by a bidirectional problem, namely learning codebook vectors $\{\mathbf{e}_i\}_{i=1}^K$ that align to the encoder outputs, and learning encoder outputs $\{z^s_q(\mathbf{x}_{\text{enc}})\}_{s=1}^{\tilde{H} \tilde{W}}$ that align to the codebook vectors \cite{snell_understanding_2021, oord_neural_2018}.
To solve this issue, the regularization term in \eqref{codebook_learning} is thus replaced by
\begin{equation}\label{SGG}
    z^s_q(\mathbf{x}_{\rm enc}) = \underset{\mathbf{e}_i}{\operatorname{argmin}} \Big \{ ||sg[\mathbf{x}^s_{\text{enc}}] - \mathbf{e}_i||^2 + \beta||\mathbf{x}^s_{\text{enc}} - sg[\mathbf{e}_i]||^2 \Big \}\quad s=1,\ldots, \tilde{H} \tilde{W},
\end{equation}
which combines the so-called codebook alignment loss, $||sg[\mathbf{x}^s_{\text{enc}}] - \mathbf{e}_i||$, with the codebook commitment loss, $||\mathbf{x}^s_{\text{enc}} - sg[\mathbf{e}_i]||$, with $sg[\cdot]$ the stop gradient operator and $\beta$ a hyperparameter to be properly tuned. The right-hand side in \eqref{SGG} implements an alternate direction minimization procedure, the stop gradient excluding the tensor it is applied to from the backpropagation.
\\
To sum up, the training of the VQ-VAE model is performed by the minimization of the loss function
\begin{equation}\label{SG}
\begin{array}{rcl}
    \mathcal{L}_{\rm VQ-VAE}(\mathbf{x}_{\text{in}}, \mathbf{x}_{\text{out}}; \mathbf{w}) & = & \mathbb{E}\big[ \lVert \mathbf{x}_{\text{in}} - \mathbf{x}_{\text{out}} \rVert^2 \big] +  ||sg[\mathbf{x}_{\text{enc}}] - z_q(\mathbf{x}_{\rm enc})||^2 
    \\[3mm]
    & + & \beta||\mathbf{x}_{\text{enc}} - sg[z_q(\mathbf{x}_{\rm enc})]||^2.
\end{array}
\end{equation}
For more details about the definition and utilization of the stop gradient operator and on VQ-VAEs, we refer the reader to \cite{oord_neural_2018}. 

%\item $||sg[\mathbf{x}_{\text{enc}}] - \mathbf{e}||_2^2$ is the codebook alignment loss. Minimizing this term, the codebook vector closest to encoder output is chosen;
%\item $||\mathbf{x}_{\text{enc}} - sg[\mathbf{e}]||_2^2$ is the codebook commitment loss. Minimizing this term, the encoder is forced to commit as much as possible to the closest codebook vector. 

\section{Unsupervised anomaly detection with convolutional autoencoders}\label{sec:leaf}
%{\color{green}fino a questo momento nessuno aveva parlato di supervised o unsupervised...non \'e per nulla chiaro poi la questione del labeled and unlabeled....
%\\
%{\color{red} Nell'introduzione si parla in maniera estensiva sia di supervised learning che di unsupervised learning. Ad un certo punto, dobbiamo stabilire a che punto fermarci nello spiegare questi concetti. Di letteratura che parla di supervised e unsupervised ce n'e a bizzeffe. Se il lettore non la conosce e vuole saperne di piu', se la va a leggere. }
%
%Come ci differenziamo dalla referenza [19]? invece in [20] dite che viene fatto un training su foglie sane MA di foglie non si parla!}
%
%\color{red} Credo che ci sia una confusione nell'uso delle numerazioni delle referenze. La referenza [20] e' soltanto una referenza alla pagina online da cui puo' essere scaricato il dataset, non e' un paper. 
%La referenza [18] si occupa di supervised classification, e noi ci differiamo da questa perche' usiamo una tecnica UNsupervised. Secondo me, dovremmo spostare la referenza [18] nell'introduzione e farla scomparire da qui. 
%
%La referenza [19] NON parla di foglie, ed e' la referenza che ho proposto di leggere a Giuseppe. Nella mail che to ho mandato un paio di giorni fa, proponevo di spostare la refernza [19] nelle cocnlusioni
%
%Da quello che ho capito io, noi siamo i primi a usare UNsupervised anomaly detection su foglie. Chi ha lavorato su foglie prima di noi, ha solo usato supervised. 
%
%Altre persone che come noi hanno usato UNSUpervised per anomaly detection, NON hanno guardato alle foglie. 
%
%}

\begin{figure}[h]
        \centering
        \includegraphics[width=10cm]{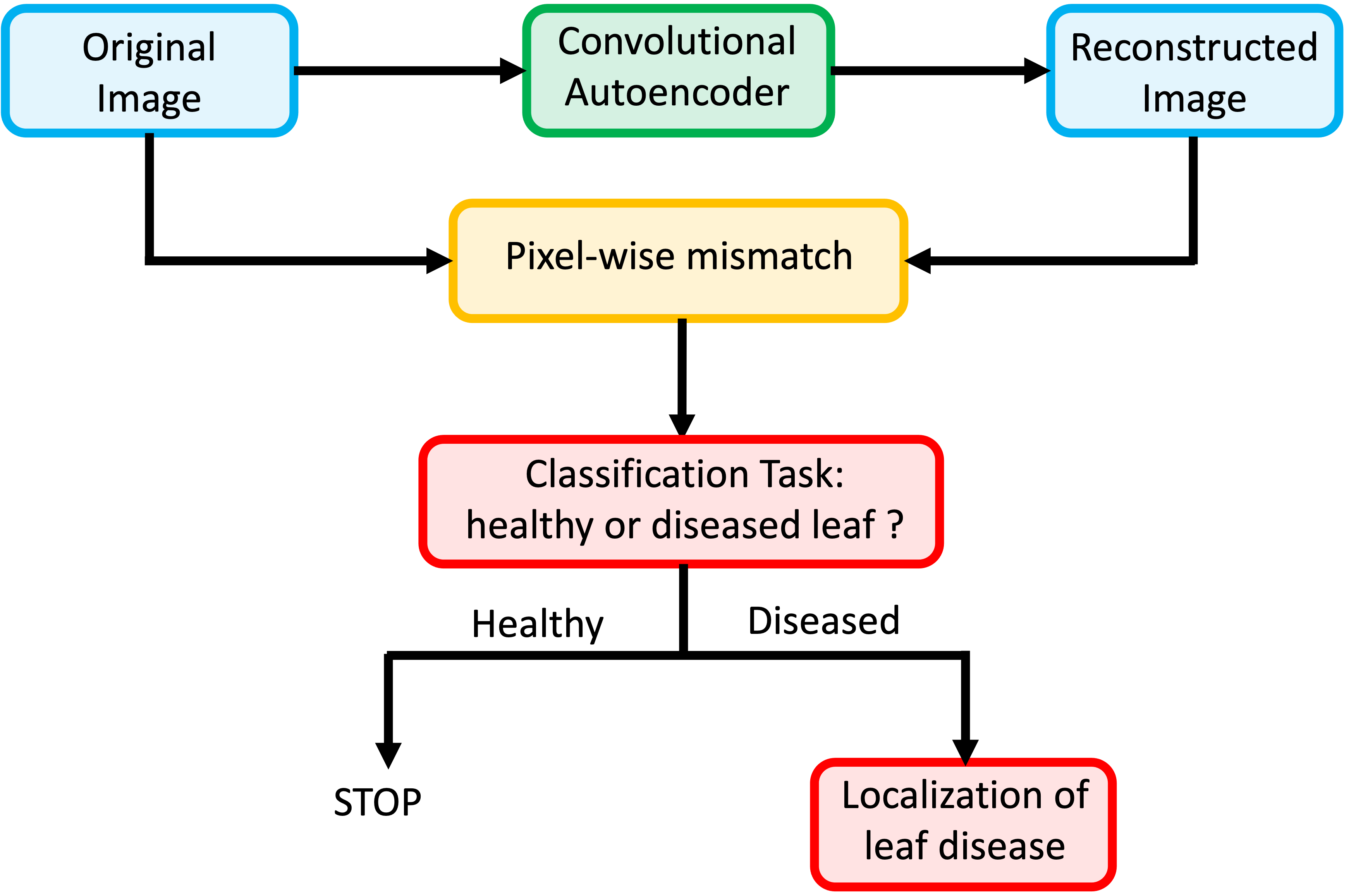}
        \caption{Schema of the proposed unsupervised DL methodology for leaf disease detection and localization.}
        \label{fig:approach-overview}
\end{figure}

Unsupervised DL-driven detection and localization of anomalies often resort to CAE models to search for patterns in input data that do not conform to images free from defects \cite{pinaya_unsupervised_2021,https://doi.org/10.48550/arxiv.1911.08616,morawski2021}
\\
To this goal, we train a CAE model only with images without anomalies. Once the CAE architecture is trained, the model is given unlabeled images. This means that the images exhibiting a defect will be reconstructed by removing the anomaly. The reconstruction error between the input and the output image is calculated and used to determine whether the image is affected by an anomaly or is not. In particular, if the reconstruction error is below a user-defined threshold, the input image is classified as free from defects. Otherwise, the image is classified as anomalous. 
Furthermore, the localization of the anomaly is carried out by analyzing the areas of the image where the pixel-wise reconstruction error is relevant (see Figure \ref{fig:approach-overview} for such a workflow particularized to a leaf dataset).

This approach has been employed to classify sets of data in diverse contexts \cite{pinaya_unsupervised_2021,morawski2021}, although, up to the authors' knowledge, not in anomaly detection of leaves in smart agriculture applications. In this section, we aim at bridging this gap by complementing the current literature, which is essentially based on supervised DL approaches \cite{9158218,10.3389/fpls.2016.01419,10.3389/fpls.2019.00941,Pintelas2021}, with an unsupervised technique.

For this task, we consider the dataset PlantVillage, which is freely available online in different versions. In more detail, we employ the dataset in the Mohanthy repository~\cite{mohanthy_plantvillage_2016}, which, in addition to the original pictures, contains preprocessed images, such as grayscale transformation and background removal.
\FloatBarrier
The dataset consists of 54314 images of healthy and diseased leaves tagged into 38 different categories based on species and illness. Table \ref{tab:plantvillage} presents the dataset composition in terms of crop and disease type, and corresponding number of images.

\begin{figure}[h]
\centering
        \begin{tabular}{cccc}
          \includegraphics[width=27mm]{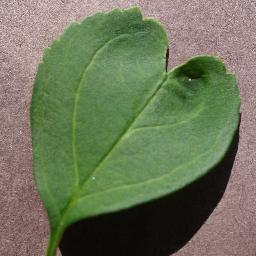} &   \includegraphics[width=27mm]{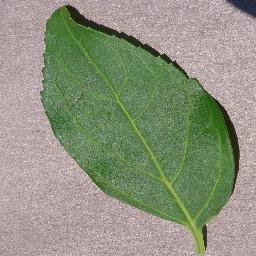} & \includegraphics[width=27mm]{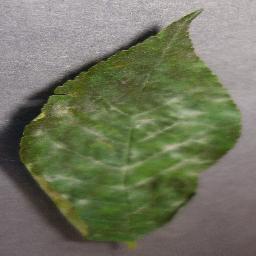} & \includegraphics[width=27mm]{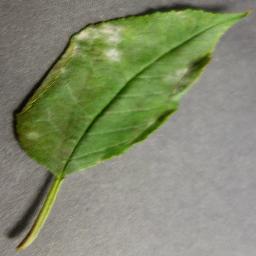}\\
        \end{tabular}
%\captionsetup{justification=centering}
        \caption{Sample images of healthy (first and second from left) and diseased (first and second from right) cherry leaves.}
        \label{cherry-samples}
    \end{figure}

\begin{figure}[h]
\centering
        \begin{tabular}{cccc}
          \includegraphics[width=27mm]{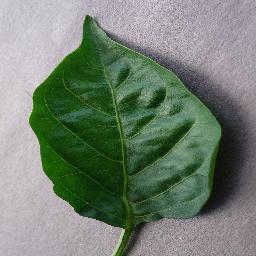} &   \includegraphics[width=27mm]{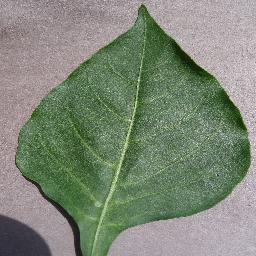} & \includegraphics[width=27mm]{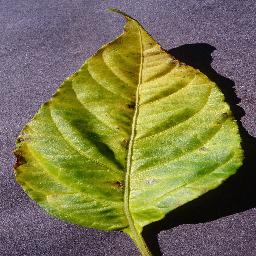} & \includegraphics[width=27mm]{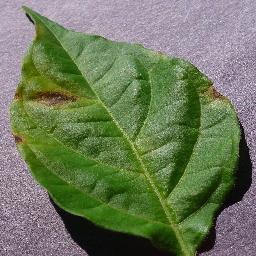}\\
        \end{tabular}
        %\captionsetup{justification=centering}
        \caption{Sample images of healthy (first and second from left) and diseased (first and second from right) pepper leaves.}
        \label{pepper-samples}
\end{figure}

Among the available species, in this work, we focus on cherry and pepper leaves (see Figures \ref{cherry-samples} and \ref{pepper-samples} for some samples).

\section{Verification}\label{sec:verification}
We numerically assess the performance of the convolutional autoencoders in Section~\ref{sec:autoencoders} when applied to the context detailed in Section~\ref{sec:leaf}. \\ A cross-comparison among CAEs, CVAEs and VQ-VAEs is carried out in terms of image reconstruction and detection and localization of leaf anomalies.
\begin{table}[h]
    \centering
    \footnotesize
    \begin{tabular}{p{3.5cm}p{3.5cm}p{4cm}}
         \hline
         \textbf{Crop} & \textbf{Disease} & \textbf{Number of images} \\
         \hline
         \hline
         \multirow{4}{*}{Apple} & Apple Scab & 630 \\
                                & Black Rot & 621 \\
                                & Cedar Apple Rust & 275 \\
                                & Healthy & 1654 \\
         \hline
         Blueberry & Healthy & 1502 \\
         \hline
         \multirow{2}{*}{Cherry} & Powdery Mildew & 1052 \\
                                 & Healthy & 854 \\
         \hline
         \multirow{4}{*}{Corn}  & Cercospora & 513 \\
                                & Common Rust & 1192 \\
                                & Nothern Leaf Blight & 1162 \\
                                & Healthy & 985 \\
         \hline
         \multirow{4}{*}{Grape}  & Black Rot & 1180 \\
                                & Esca & 1383 \\
                                & Blight & 1076 \\
                                & Healthy & 423 \\
         \hline
         Orange  & Haunglongbing & 5507 \\
         \hline
         \multirow{2}{*}{Peach}  & Bacterial Spot & 2297 \\
                                & Healthy & 360 \\
         \hline
         \multirow{2}{*}{Pepper}  & Bacterial Spot & 997 \\
                                & Healthy & 1478\\
         \hline
         \multirow{3}{*}{Potato}  & Early Blight & 1000 \\
                                & Late Blight & 1000 \\
                                & Healthy & 152 \\
         \hline
         Raspberry & Healthy & 371 \\
         \hline
         Soybean & Healthy & 5090 \\
         \hline
         Squash & Powdery Mildew & 1835 \\
         \hline
         \multirow{2}{*}{Strawberry}  & Leaf Scorch & 1109 \\
                                & Healthy & 456 \\
         \hline
         \multirow{10}{*}{Tomato}  & Bacterial Spot & 2127 \\
                                & Early Blight & 1000 \\
                                & Late Blight & 1909 \\
                                & Leaf Mold & 952 \\
                                & Septoria Leaf Spot & 1771 \\
                                & Spider Mites & 1676 \\
                                & Target Spot & 1404 \\
                                & Yellow Leaf Curl Virus & 373 \\
                                & Mosaic Virus & 5357 \\
                                & Healthy & 1591 \\
         \hline
         \hline
    \end{tabular}
    %\captionsetup{justification=centering}
    \caption{PlantVillage dataset composition.}
    \label{tab:plantvillage}
\end{table}
\FloatBarrier
\FloatBarrier
\subsection{Data preprocessing}
Data preprocessing is a fundamental phase of many machine learning and data mining approaches which manipulates, deletes and integrates the original data in order to improve the performance of the selected model.

Concerning the cherry and pepper datasets selected in the previous section, we essentially resort to resizing and data augmentation \cite{data_augmentation}.
\\
Since we perform an accelerated training by GPU hardware, we resize each input RGB image to $256 \times 256$ pixels in order to cope with memory limitations of the GPU device.
Data augmentation is employed since DL models require large volumes of data for accuracy reasons. To increase the number of images available for training, we resort to the following transformations on the original data:

\begin{itemize}
    \item image flipping: an image is mirrored around its horizontal or vertical axis\footnote{The image flipping around the vertical axis is also known as flopping};
    \item image rotation: an image is rotated in a clockwise or counterclockwise direction given a rotation angle $\theta$. Since the models accept only squared images, the possible rotation angles are $90^{\circ}$, $180^{\circ}$ and $270^{\circ}$.
\end{itemize}

The two datasets are split into 3 different subsets, to be used for training, validation and testing (see Table \ref{tab:data_split} for more details).
The training dataset is composed by only healthy leaves images, which are used to optimize the DL models by minimizing the loss functions in \eqref{MSE}, \eqref{KL}, and \eqref{SG}, respectively. The validation is successively adopted to identify the best model among CAE, CVAE e VQ-VAE in terms of reconstruction of both healthy and unhealthy leaf images.
Finally, the testing dataset consists of pictures of the two classes of cherry and pepper leaves, the goal of this phase being the detection and the localization of leaf anomalies.
\begin{table}[h!]
        \centering
        \begin{tabular}{|c|c|c|c|c|}
            \hline
              & Training set & Validation set & \multicolumn{2}{c|}{Test set} \\
             \cline{2-5}
              & Healthy & Healthy & Healthy & Diseased \\
             \hline
             \hline
             cherry & 681 & 85 & 85 & 85 \\
             pepper & 1134 & 113 & 113 & 113 \\
             \hline
        \end{tabular}
        \captionsetup{justification=centering}
        \caption{Dataset splitting used for training, validation and test.}
        \label{tab:data_split}
\end{table}

\subsection{Model setup}

We provide the architectural and training setup of the CAE, CVAE and VQ-VAE models. In particular,
the three networks share the same architecture for the encoder and decoder parts, while being characterized by a different latent space configuration.

\subsubsection{CAE architecture and training setup}
The CAE takes in input an image with size $(H, W, C)$ $= (256, 256, 3)$ and passes it to the encoder.
The encoder is composed by $5$ convolutional layers and $5$ max-pooling layers for downsampling, that return a tensor of size $(4, 4, 64)$.
\\
The downsampled tensor is then passed to the layers that define the latent space. The first one is a flatten layer that transforms the three-dimensional tensor into a vector ${\bf x}_{\rm enc}$ with $1024$ entries. The encoded vector is fed to two fully connected layers, each of them reducing the length of the vector by a factor of 8, so that the vector eventually matches the size of the latent space with 16 entries. In reverse, another pair of fully connected layers increases the size of the vector, each by a factor of 8, yielding the vector ${\bf x}_{\rm lat}$ matching the original size (i.e., $1024$ entries). 
The vector is then passed to a view layer that reshapes it into a three-dimensional tensor of size $(4, 4, 64)$ and then fed to the decoder, which consists of $5$ convolutional and of $5$ upsampling layers. The output ${\bf x}_{\rm out}$ with size $(H, W, C) = (256, 256, 3)$ is finally generated.
\\
The ReLU activation function is used for all the convolutional and deconvolutional layers, except for the output layer, where the sigmoid activation function is employed to ensure that the pixel values are within the range $[0,1]$.

The CAE training is performed over $200$ epochs by resorting to Adam algorithm \cite{kingma_adam_2017}, with a learning rate of $10^{-3}$ for the optimization.

\subsubsection{CVAE architecture and training setup}
Since the encoder of the CVAE resorts to a probability distribution to describe the input data, the latent space performs the so-called reparametrization trick in order to generate a sample from such a distribution to be eventually passed to the decoder. 
\\
%The reparametrization trick function returns the sampled tensor 
%\begin{equation}
%    \mathbf{x}_{\rm lat} = \mathbf{x}_{\mu} + \exp(\eta \mathbf{x}_{\sigma}),
%    \label{eq:reparam}
%\end{equation}
%{\color{green} xmu o muenc? analogamente xsigma o Sigmaenc come in Fig 2?}
%with $\eta$ a parameter to be properly selected. Here, we set $\eta$ to $0.1$ as a result of a grid search hyperparameter optimization.
%\\
The CVAE decoder has the same architecture as for the CAE model, and starting from $\mathbf{x}_{\rm lat}$ reconstructs the new image $\mathbf{x}_{\rm out}$ with size $(H, W, C) = (256, 256, 3)$.

The training is performed with the same optimizer and learning rate as for the CAE, although relying only on $100$ epochs due to the increased computational complexity of the DL model.

\subsubsection{VQ-VAE architecture and training setup}
The input image, with size $(H, W, C)$ $= (256, 256, 3)$, is passed to the encoder which is composed by 3 convolutional layers without any max-pooling layer for downsampling. A tensor of size $(4, 4, 64)$ is thus produced. Each thread of size $(1,1,64)$ of the tensor is replaced by the closest vector of the codebook for quantization. The codebook size is $(K, \tilde{C}) = (512, 64)$, meaning that the discrete space of the codebook is made of 512 vectors, each with 64 entries.
\\
The quantization of the latent space yields a tensor with size $(4, 4, 64)$ which is provided as an input to the decoder constituted by 3 convolutional layers without any upsampling layer.

The training phase is performed with the same optimizer and learning rate as for the previous two DL models, by resorting to only $50$ epochs due to a further increase in the computational cost of the VQ-VAE architecture.

\subsection{Hardware \& Software Specifics}

The numerical results presented in this work are obtained with the hardware and OS specifics in Table~\ref{tab:hw_sw}.

\begin{table}[h!]
    \centering
    \begin{tabular}{|c|c|}
        \hline
        \textbf{PC model} & Lenovo ThinkPad P14s \\
        \hline
        \textbf{CPU} & Intel® Core™ i7-10510U CPU @ 1.80GHz × 8 \\
        \hline
        \textbf{RAM} & 16 GB DDR4-3200 \\
        \hline
        \textbf{GPU} & NV138 / Mesa Intel® UHD Graphics (Integrated) \\
        \hline
        \textbf{OS} & Ubuntu 20.04 LTS 64bit \\
        \hline
    \end{tabular}
    \captionsetup{justification=centering}
    \caption{Hardware and software specifications.}
    \label{tab:hw_sw}
\end{table}

The DL models are built and trained using the well-known neural network library \texttt{Pytorch} \cite{pytorch2019, pytorch}, which enables a seamless remapping of the tensors from CPUs to GPUs for hardware accelerated training.

\section{Discussion}\label{sec:discussion}

In this section, we compare the three considered autoencoders in terms of reconstruction and classification of both healthy and anomalous leaves, as well as disease localization. These three skills are mutually related. Indeed, the capability of a model to accurately reproduce a non-anomalous image and to clean the defects on the diseased leaves is crucial in view of a reliable classification and localization.

\subsection{Reconstruction and anomaly removal performance}
\label{section:recon}
We compare the predictive performance of the CAE, CVAE, VQ-VAE models to reconstruct anomalous and normal samples in the test portion of the dataset (see Table \ref{tab:data_split}) for cherry and pepper leaf images. The metric used to quantify the accuracy of the reconstruction is the MSE,
%Since the three models take different times to train due to the specific complexity, the comparison in terms of accuracy is performed both as a function of the number of epochs and of the computational time for training. 
whose values - scaled by a $10^3$ factor - are collected in Table~\ref{tab:validation-mse-cherry} and Table~\ref{tab:validation-mse-pepper} (second-fourth column).
The discrepancy $\Delta$ between the MSE values for healthy and diseased samples is also provided in both tables.
\\
%Due to the stochasticity of the training, the accuracy is not necessarily monotonically increasing with respect to the number of epochs. Therefore, the MSE value in the table is the one associated with the best performing epoch whose index is highlighted in brackets.
As expected by the architecture of the three models, we can rank VQ-VAE as the best performing architecture, followed by CAE and CVAE, independently of the considered leaf species.
Moreover, we can appreciate how the VQ-VAE yields larger values for $\Delta$. This emphasizes the separation of the two classes, with a consequent improvement in terms of leaf classification.
\begin{table}[h!]
    \centering
    \begin{tabular}{|c|c|c|c||c|c|}
        \hline
         dataset & CAE & CVAE & VQ-VAE & CAE-TE & CVAE-TE  \\
         \hline
         \hline
         healthy & 2.1766 & 4.8256 & 1.2380 & 2.1577 & 4.6332 \\
         diseased & 2.1988 & 5.0732 & 2.0052 & 2.1739 & 4.6441 \\
         $\Delta$ & 0.0222 & 0.2476 & 0.7672 & 0.0162 & 0.0109 \\
         \hline
    \end{tabular}
    %\captionsetup{justification=centering}
    \caption{Reconstruction accuracy for the cherry dataset: comparison among CAE, CVAE and VQ-VAE in terms of MSE.}
    \label{tab:validation-mse-cherry}
\end{table}
\begin{table}[h!]
    \centering
    \begin{tabular}{|c|c|c|c||c|c|}
        \hline
         dataset & CAE & CVAE & VQ-VAE & CAE-TE & CVAE-TE  \\
         \hline
         \hline
         healthy & 4.1029 & 6.2846 & 1.1674 & 3.9961 & 6.1512 \\
         diseased & 4.7811 & 6.3255  & 1.8971 & 4.4350 & 6.2846 \\
         $\Delta$ & 0.6782 & 0.0409 & 0.7297 & 0.4389 & 0.1334 \\
         \hline
    \end{tabular}
    %\captionsetup{justification=centering}
    \caption{Reconstruction accuracy for the pepper dataset: comparison among CAE, CVAE and VQ-VAE in terms of MSE.}
    \label{tab:validation-mse-pepper}
\end{table}

Since CAEs, CVAEs, VQ-VAEs take different times to train due to the specific complexity, we carry out a further comparison in order to consider such a mismatch. 
On average, VQ-VAE takes $4$ times the training time required by CAE and CVAE, independently of the considered dataset. This is highlighted in Table \ref{tab:trainin_time} that shows the total training time (in minutes) measured as the median over $10$ different runs for the three models. 
\begin{table}[h!]
    \centering
    \begin{tabular}{|c|c|c|c|}
        \hline
        dataset  & CAE & CVAE & VQ-VAE \\
        \hline
        \hline
         cherry & 43 & 49 &188 \\
         \hline
         pepper & 56 & 61 & 220 \\
        \hline
    \end{tabular}
    \caption{Training time in minutes for cherry and pepper datasets.}
    \label{tab:trainin_time}
\end{table}
\\
Thus, the three DL networks are compared for a fixed computational training time, TE, that we select as the time required by VQ-VAE to complete $50$ epochs (fourth column in Table~\ref{tab:trainin_time}). The MSE associated with this investigation is provided in Table~\ref{tab:validation-mse-cherry} and Table~\ref{tab:validation-mse-pepper} (fifth-sixth column). Despite the increased number of exploited epochs, both CAE and CVAE slightly improve the accuracy, while VQ-VAE remains the best performing model both in terms of reconstruction and of class separation.

To corroborate the performed investigation from a qualitative viewpoint, we consider the $4$ images of healthy and unhealthy pepper and cherry leaves in Table \ref{tab:recon-results} (first row).
\\
When images of healthy leaves are provided as an input, the autoencoders are expected to reproduce the original image as accurately as possible. Although CAE and CVAE can recover the main characteristics of the leaf, such as color and shape, both the models are unable to reproduce the details at the fine scale, such as veins, chromatic differences and edges (second and third row, first and third column).
On the contrary, the VQ-VAE model reproduces the images at a higher resolution, thus outperforming the other two architectures. In fact, now the chromatic differences of the veins of the leaves as well as the detail of the edges are clearly visible in the reconstructed image (fourth row, first and third column). 

When images of unhealthy leaves are provided as an input, the autoencoder is expected to maintain the shape of the leaf while cleaning the surface from the alterations due to the disease (anomaly removal).
We notice that the CAE and CVAE architectures mildly alter the shape of the leaf and lose each detail at a fine scale (second and third row, second and fourth column).
%Moreover, we recognize a different quality reconstruction when considering cherry or pepper leaves, likely due to the different features exhibited by the disease.
%In the cherry dataset case, the defect covers the whole leaf surface so that the reconstruction turns out to be very uniform; on the contrary, the pepper dataset is characterized by pointwise anomalies and the associated reconstruction results in a heterogeneous texture.
On the contrary, VQ-VAE architecture faithfully reproduces the shape of the leaf as well as the details at a fine scale.
Besides being more accurate, the VQ-VAE model has also enhanced generalizability properties. 
While accuracy and generalizability are both strongly desired in an autoencoder for image reconstruction, generalizability may actually be counterproductive in the context of anomaly removal. 
In fact, an effective generalizable autoencoder may accurately reconstruct the anomaly along with all the other features of the input image. This issue is acknowledged also by other works in the literature, especially for data where the anomaly covers a very small surface of the image \cite{gong_memorizing_2019}.
\\
The generalizability phenomenon is partially observed in the results in Table \ref{tab:recon-results} (fourth row, second and fourth column), where the reconstructed images exhibit a certain heterogeneity in the color distribution. As a matter of fact, the MSE error for the diseased leaves in Tables~\ref{tab:validation-mse-cherry} and~\ref{tab:validation-mse-pepper} is lower for the VQ-VAE with respect to the two other architectures.

To sum up, we can conclude that the VQ-VAE represents the best performing model as far as image reconstruction and anomaly removal are concerned.
\begin{table}[h!]
    \centering
    \begin{tabular}{|c|c|c|c|c|}
        \hline
         & cherry healthy & cherry diseased & pepper healthy & pepper diseased \\
        \hline
        \hline
        \rotatebox[origin=c]{90}{original} &
        \makecell{\includegraphics[width=27mm,height=27mm,keepaspectratio]{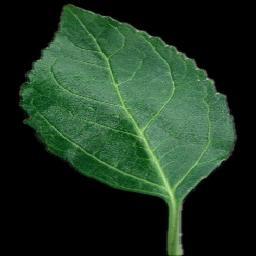}} &
        \makecell{\includegraphics[width=27mm,height=27mm,keepaspectratio]{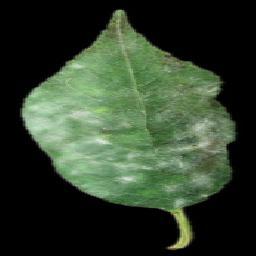}} &
        \makecell{\includegraphics[width=27mm,height=27mm,keepaspectratio]{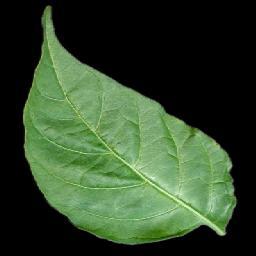}} &
        \makecell{\includegraphics[width=27mm,height=27mm,keepaspectratio]{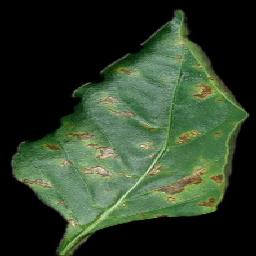}} \\
        \hline
        \rotatebox[origin=c]{90}{CAE} &
        \makecell{\includegraphics[width=27mm,height=27mm,keepaspectratio]{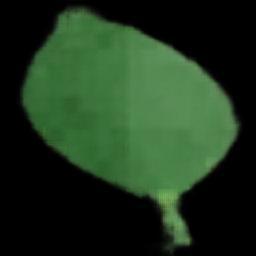}} &
        \makecell{\includegraphics[width=27mm,height=27mm,keepaspectratio]{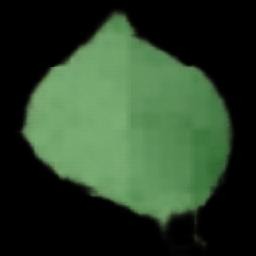}} &
        \makecell{\includegraphics[width=27mm,height=27mm,keepaspectratio]{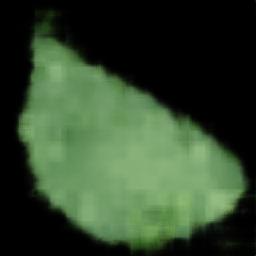}} &
        \makecell{\includegraphics[width=27mm,height=27mm,keepaspectratio]{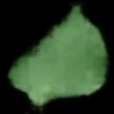}} \\
        \hline
        \rotatebox[origin=c]{90}{CVAE} &
        \makecell{\includegraphics[width=27mm,height=27mm,keepaspectratio]{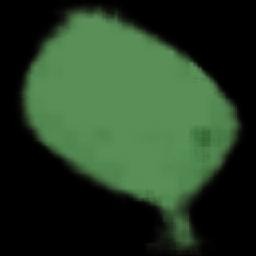}} &
        \makecell{\includegraphics[width=27mm,height=27mm,keepaspectratio]{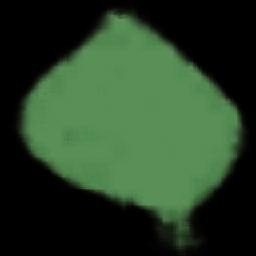}} &
        \makecell{\includegraphics[width=27mm,height=27mm,keepaspectratio]{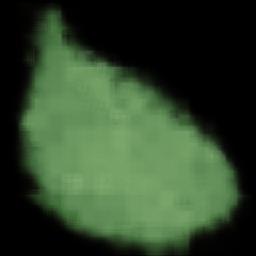}} &
        \makecell{\includegraphics[width=27mm,height=27mm,keepaspectratio]{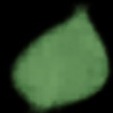}} \\
        \hline
        \rotatebox[origin=c]{90}{VQ-VAE} &
        \makecell{\includegraphics[width=27mm,height=27mm,keepaspectratio]{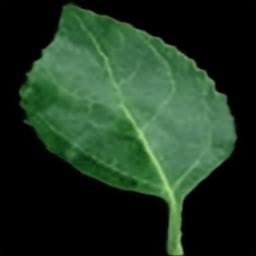}} &
        \makecell{\includegraphics[width=27mm,height=27mm,keepaspectratio]{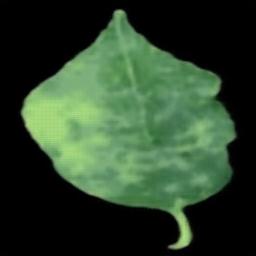}} &
        \makecell{\includegraphics[width=27mm,height=27mm,keepaspectratio]{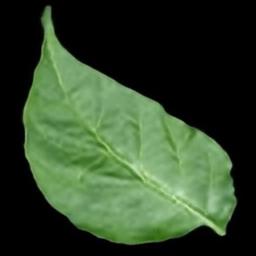}} &
        \makecell{\includegraphics[width=27mm,height=27mm,keepaspectratio]{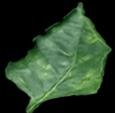}} \\
        \hline
    \end{tabular}
    %\captionsetup{justification=centering}
    \caption{Reconstruction and anomaly removal: comparison among CAE, CVAE and VQ-VAE on both cherry and pepper healthy and diseased leaves.}
    \label{tab:recon-results}
\end{table}

\subsection{Leaf anomaly detection}

In this section, we address the classification between healthy and unhealthy leaves by resorting to anomaly detection.
In particular, to quantify the detection capability characterizing the considered unsupervised architectures, we resort to the AUC-ROC score, so that higher values of such a quantity reveal a better classification of the model. We observe that an accurate classification strictly depends on the reconstruction property of the selected DL method.
\\
In Table \ref{tab:classification-results} we provide the values of the AUC-ROC score for the three considered models. 
\begin{table}[h!]
    \centering
    \begin{tabular}{|c|c|c|c||c|c|}
        \hline
         dataset & CAE & CVAE & VQ-VAE & CAE-TE & CVAE-TE \\
         \hline
         \hline
         cherry & 0.965 & 0.946 & 0.983
         & 0.968 & 0.941 \\
         pepper & 0.892 & 0.819 & 0.936
         & 0.888 & 0.844 \\
         \hline
    \end{tabular}
    %\captionsetup{justification=centering}
    \caption{Anomaly detection: comparison among CAE, CVAE and VQ-VAE in terms of AUC-ROC score.}
    \label{tab:classification-results}
\end{table}
CAE and CVAE architectures better distinguish healthy leaves from unhealthy ones on the cherry dataset than on the pepper dataset (second-third columns). This is likely due to different anomalies of the diseases that affect the two types of plants.
In particular, the uniform distribution of the disease over the whole cherry leaf facilitates a clear distinction between the input (striped) unhealthy and the reconstructed (completely cleaned) images. This results into a good separability of the healthy and unhealthy classes and enables a reliable classification when using CAEs and CVAEs.

In the previous section, VQ-VAE proved to be characterized by enhanced generalizability properties. This is confirmed by the MSE values in Tables~\ref{tab:validation-mse-cherry} and~\ref{tab:validation-mse-pepper} which are lower for the VQ-VAE when compared with CAE and CVAE ($2.50$ and $3.33$ times lower for the pepper dataset). Despite the superior generalizability feature, VQ-VAE outperforms CAEs and CVAEs in terms of leaf classification due to the highest separation capability, corroborated by the values of the discrepancy $\Delta$ in the tables.

Also in the anomaly detection context, we perform a cross-comparison among the three models for a time equivalent training phase. The value of the AUC-ROC score associated with this analysis is provided in the last two columns of Table~\ref{tab:classification-results}. Similarly to the reconstruction and anomaly removal verification, we can remark that the performance of both CAE-TE and CVAE-TE models do not significantly change by varying the number of epochs. Also in this investigation, VQ-VAE still represents the most reliable method.

\subsection{Leaf anomaly localization}

As a last check, we assess the anomaly localization capability of the CAE, CVAE and VQ-VAE models. With this regard, the performance is a direct consequence of the reconstruction properties discussed in Section~\ref{section:recon}, the localization step being carried out moving from the distribution of the reconstruction error. 
\begin{table}[h!]
    \centering
    \begin{tabular}{|c|c|c|c|c|}
        \hline
         & cherry healthy & cherry diseased & pepper healthy & pepper diseased \\
        \hline
        \hline
        \rotatebox[origin=c]{90}{Original} &
        \makecell{\includegraphics[width=27mm,height=27mm,keepaspectratio]{images/results/or-cherry-h-1.jpg}} &
        \makecell{\includegraphics[width=27mm,height=27mm,keepaspectratio]{images/results/or-cherry-d-1.jpg}} &
        \makecell{\includegraphics[width=27mm,height=27mm,keepaspectratio]{images/results/or-pepper-h-1.jpg}} &
        \makecell{\includegraphics[width=27mm,height=27mm,keepaspectratio]{images/results/or-pepper-d-2.jpg}} \\
        \hline
        \rotatebox[origin=c]{90}{CAE} &
        \makecell{\includegraphics[width=27mm,height=27mm,keepaspectratio]{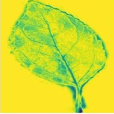}} &
        \makecell{\includegraphics[width=27mm,height=27mm,keepaspectratio]{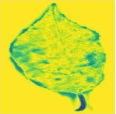}} &
        \makecell{\includegraphics[width=27mm,height=27mm,keepaspectratio]{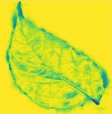}} &
        \makecell{\includegraphics[width=27mm,height=27mm,keepaspectratio]{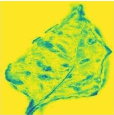}} \\
        \hline
        \rotatebox[origin=c]{90}{CVAE} &
        \makecell{\includegraphics[width=27mm,height=27mm,keepaspectratio]{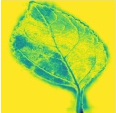}} &
        \makecell{\includegraphics[width=27mm,height=27mm,keepaspectratio]{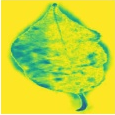}} &
        \makecell{\includegraphics[width=27mm,height=27mm,keepaspectratio]{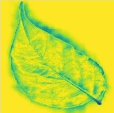}} &
        \makecell{\includegraphics[width=27mm,height=27mm,keepaspectratio]{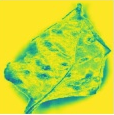}} \\
        \hline
        \rotatebox[origin=c]{90}{VQ-VAE} &
        \makecell{\includegraphics[width=27mm,height=27mm,keepaspectratio]{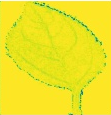}} &
        \makecell{\includegraphics[width=27mm,height=27mm,keepaspectratio]{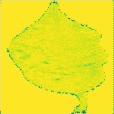}} &
        \makecell{\includegraphics[width=27mm,height=27mm,keepaspectratio]{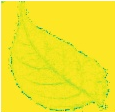}} &
        \makecell{\includegraphics[width=27mm,height=27mm,keepaspectratio]{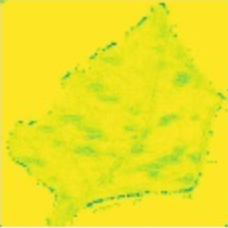}} \\
        \hline
    \end{tabular}
    %\captionsetup{justification=centering}
    \caption{Anomaly localization: comparison in terms of reconstruction error among CAE, CVAE and VQ-VAE on both cherry and pepper healthy and diseased leaves.}
    \label{tab:local-results}
\end{table}

\begin{figure}[h!]
    \centering
        \includegraphics[width=27mm,height=27mm,keepaspectratio]{images/results/or-cherry-d-1.jpg}
        \includegraphics[width=27mm,height=27mm,keepaspectratio]{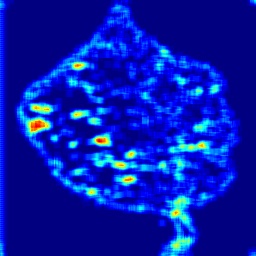}
        \includegraphics[width=27mm,height=27mm,keepaspectratio]{images/results/or-pepper-d-2.jpg}
        \includegraphics[width=27mm,height=27mm,keepaspectratio]{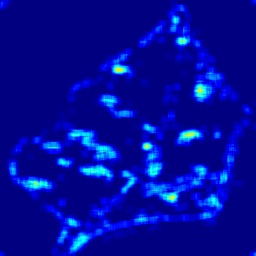}
    %\captionsetup{justification=centering}
    \caption{Anomaly localization: comparison between the original image and the high-contrast reconstruction error for the VQ-VAE model, cherry (first and second from left) and pepper (first and second from right) leaves.}
    \label{tab:post-processing}
\end{figure}

In Table~\ref{tab:local-results}, we show such an error for both the healthy and unhealthy leaf samples in Table~\ref{tab:recon-results}. We adopt a yellow-to-blue color map, after normalizing the images so that the reconstruction error ranges in $[0,1]$.
The anomaly localization provided by CAE and CVAE turns out to be rather unpractical. Indeed, all the details (such as, veins, edges, chromatic differences) at the fine scale which are missed during the reconstruction are highlighted as an error, analogously to the actual anomalies. This undesirable behavior is particularly evident in the CVAE outputs.
On the contrary, the high quality guaranteed by VQ-VAE in sharply reconstructing the small-scale features and in mildly recovering the anomalies leads to confine the error to the edges and to the veins for the healthy samples, and to the actual defects for the diseased leaves (we refer to Figure~\ref{tab:post-processing} for a high-contrast post-processing of the VQ-VAE reconstruction error for the diseased cherry and pepper leaves in Table~\ref{tab:local-results}).

These considerations allow us to conclude that, when applied to PlantVillage dataset, VQ-VAE models perform better with respect to CAE and CVAE, also in terms of anomaly localization.

\section{Conclusions and future work}

The employment of unsupervised DL techniques on the PlantVillage dataset allowed us to compare the performance of standard autoencoders, such as CAE and CVAE model, with the more sophisticated VQ-VAE architecture, when applied to healthy and unhealthy leaves. The investigation focused on $2$ out of the $14$ different plant species included in the dataset, namely cherry and pepper leaves affected by powdery mildew and bacterial spot, respectively. A cross-comparison among the three autoencoders has been carried out in terms of: i) image reconstruction for both normal and anomalous samples, ii) anomaly removal from diseased leaves, iii) anomaly detection, iv) anomaly localization.

The results in Section~\ref{sec:discussion} show that VQ-VAEs have a superior predictive performance than CAEs and CVAEs in the autoencoding task (see Table \ref{tab:recon-results}), which also reflects into a more accurate image classification of healthy and unhealthy leaves (see Table \ref{tab:classification-results}). The improved generalizability of the VQ-VAE models leads to a significant reduction of the reconstruction error on every image (see Tables \ref{tab:validation-mse-cherry}, \ref{tab:validation-mse-pepper} and \ref{tab:local-results}). This might potentially be counterproductive in the context of anomaly detection because it may reduce the capability of the model to separate images belonging to different classes. This possible risk has already been addressed in other contexts, such as \cite{gong_memorizing_2019}. However, for the specific dataset studied in this work, the gap between the reconstruction errors of the VQ-VAE on healthy and unhealthy images is significantly wider than for CAEs and CVAEs (see Tables \ref{tab:validation-mse-cherry} and \ref{tab:validation-mse-pepper}), thus allowing the VQ-VAE to retain a superior capacity in accurately separating images of healthy leaves from images of unhealthy leaves. 

Future work will be dedicated to apply methodology recently proposed in other contexts~\cite{10.1145/3447548.3467417} to plant anomaly detection. These approaches combine supervised and semi-supervised DL methods to address situations where readily accessible large-scale unlabeled data may contain both known and unknown anomalies,
\\
As a further development of interest, we consider the issue of miss detection of the anomaly due to a highly generalizable autoencoder that accurately reconstructs both normal and anomalous data. The outcome of this future effort will increase the robustness of current DL techniques for anomaly detection.

\section*{Acknowledgments}
Massimiliano Lupo Pasini thanks Dr. Vladimir Protopopescu for his valuable feedback in the preparation of this manuscript.
Massimiliano Lupo Pasini's work was supported in part by the Artificial Intelligence Initiative as part of the Laboratory Directed Research and Development (LDRD) Program of Oak Ridge National Laboratory, managed by UT-Battelle, LLC, for the US Department of Energy under contract DE-AC05-00OR22725.
\\
Simona Perotto gratefully acknowledges the financial support of INdAM - GNCS Project 2022 ``Metodi di riduzione computazionale per le scienze applicate: focus su sistemi complessi''.

\bibliographystyle{abbrv}
\bibliography{references}

\end{document}